%% file: acl2023.tex
\pdfoutput=1

\documentclass[11pt]{article}

\usepackage[]{ACL2023}

\usepackage{times}
\usepackage{latexsym}
\usepackage{graphicx}
\usepackage[T1]{fontenc}
\usepackage{subfigure}
\usepackage{tabularx,booktabs}
\usepackage[utf8]{inputenc}
\usepackage{enumitem}
\usepackage{microtype}
\usepackage{amsmath}
\usepackage{inconsolata}
\usepackage{amssymb}
\usepackage{multirow}

\title{Continual Dialogue State Tracking via Reason-of-Select Distillation}

\author{Yujie Feng$^1$, Bo Liu$^1$, Xiaoyu Dong$^1$, Zexin Lu$^1$ \\ \textbf{Li-Ming Zhan}$^1$, \textbf{Albert Y.S. Lam}$^2$, \textbf{Xiao-Ming Wu}$^1$\thanks{ ~ Corresponding author.} \\
$^1$Department of Computing, The Hong Kong Polytechnic University, Hong Kong S.A.R.\\
$^2$Fano Labs, Hong Kong S.A.R.\\
 yujie.feng@connect.polyu.hk, xiao-ming.wu@polyu.edu.hk 
}

\begin{document}
\maketitle
\begin{abstract}
\input{0.abstract}

\end{abstract}

\section{Introduction}
\input{1.introduction}

\section{Motivation for Boosting Meta-reasoning Capability in Continual DST}
\input{2.motivation}

\section{Reason-of-Select Distillation}
\input{3.method}

\section{Experiments}\label{sec:exp}
\input{4.exp}

\section{Related Work}
\input{5.related}

\section{Conclusion}
\input{6.conclusion}

\section*{Limitations}
\input{7.limitations}

\section*{Acknowledgments}
We thank the anonymous reviewers for their valuable feedback. This research was partially supported by the grant of HK ITF ITS/359/21FP.

\bibliography{acl2023}
\bibliographystyle{acl_natbib}

\appendix
\input{8.appendix}

\end{document}

%% file: 0.abstract.tex
An ideal dialogue system requires continuous skill acquisition and adaptation to new tasks while retaining prior knowledge. 
Dialogue State Tracking (DST), vital in these systems, often involves learning new services and confronting catastrophic forgetting, along with a critical capability loss termed the ``Value Selection Quandary.'' 
To address these challenges, we introduce the Reason-of-Select (RoS) distillation method by enhancing smaller models with a novel `meta-reasoning' capability. 
Meta-reasoning employs an enhanced multi-domain perspective, combining fragments of meta-knowledge from domain-specific dialogues during continual learning. This transcends traditional single-perspective reasoning. The domain bootstrapping process enhances the model's ability to dissect intricate dialogues from multiple possible values. Its domain-agnostic property aligns data distribution across different domains, effectively mitigating forgetting. 
Additionally, two novel improvements, ``multi-value resolution'' strategy and Semantic Contrastive Reasoning Selection method, significantly enhance RoS by generating DST-specific selection chains and mitigating hallucinations in teachers' reasoning, ensuring effective and reliable knowledge transfer.
Extensive experiments validate the exceptional performance and robust generalization capabilities of our method.
The source code\footnote{\url{https://github.com/WoodScene/RoS}} is provided for reproducibility.




%% file: 1.introduction.tex
Practical dialogue systems require continual adaptation to new services while maintaining previous task capabilities. However, previous research in dialogue systems has focused on domain-specific offline systems, lacking adaptation abilities~\cite{ni2023recent}. 
With recent advances in large language models (LLMs), LLM-based systems have shown significant superiority over previous methods~\cite{hu2023enhancing}.
However, the bulky sizes of LLMs make retraining models from scratch prohibitively time-consuming and challenging~\cite{liu2023good}.
Thus, efficient continual learning (CL) is vital for dialogue systems to obtain new skills while retaining knowledge of previous tasks. 
Dialog State Tracking (DST), central to task-oriented dialogue systems, dynamically updates triplets (domain, slot, value) to manage user intents~\cite{feng2023towards}. The growing necessity to expand DST models for new services has spurred interest in recent Continual DST task~\cite{cho2023continual}.

\begin{figure}[t]
  \centering
  \includegraphics[width=1.0\linewidth]{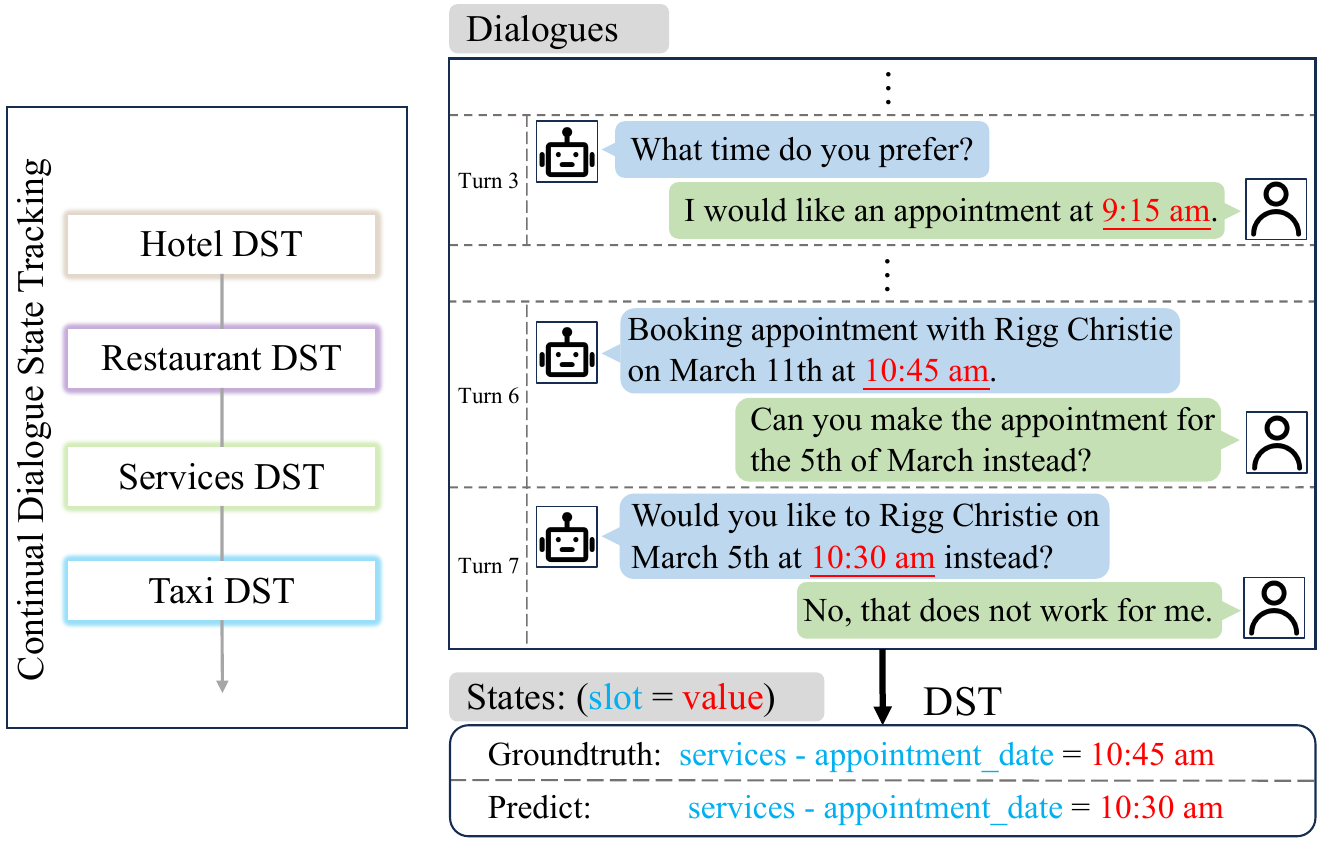}
  \caption{\textit{Left:} Depiction of the Continual DST learning process. \textit{Right:} An actual instance of the ``Value Selection Quandary'' phenomenon, demonstrating a dialogue with three mentioned date values, where the model incorrectly chooses the most recent time at turn 7 rather than the correct value at turn 6.}
  \label{fig:intro}
\end{figure}

As a specific task of CL, Continual DST grapples with catastrophic forgetting~\cite{mccloskey1989catastrophic, french1999catastrophic}, where sequential learning of new tasks hinders retention of prior ones. This problem is more acute in DST due to significant data distribution shifts across domains in the CL process, exemplified by transitions from Hotel to Restaurant domains (see Figure~\ref{fig:intro}). Recent advances in Continual DST, including memory replay and regularization \cite{zhu2022continual}, and reformulation as a question-answering task \cite{cho2023continual}, strive to address forgetting. However, these methods face challenges like reliance on past data and high computational demands during testing, which hinder real-time applications.

Recent LLMs show impressive DST performance \cite{heck2023chatgpt, feng2023towards}, but practical deployment faces hurdles like offline computational load and online data privacy concerns. Moreover, enabling LLMs for CL demands significant resources, leading to the exploration of smaller models.
After a comprehensive analysis of the current smaller DST models, we have identified a critical loss of capability brought by domain shifts in CL that models fail to output correct value when facing a set of similar candidates (named ``Value Selection Quandary'' as elaborated in Section 2).
For example, in Figure~\ref{fig:intro}, when tracking the <services-appointment\_date> slot, the model incorrectly chooses the most recent time mentioned, revealing its inability to grasp contextual subtleties and favoring direct value extraction over logical reasoning. 
This phenomenon is because, as the CL progresses, the model needs to understand the relevant knowledge of newly emerging domains and chooses to forget domain-irrelevant or weakly relevant knowledge, such as value selection.

This paper addresses these challenges by boosting smaller models' reasoning ability, termed as \textbf{\emph{meta-reasoning}}. 
Meta-reasoning, guided by the insight that domain-specific dialogues represent only a fragment of the underlying meta-knowledge, can be viewed as a form of multi-view augmentation from different domains in the CL process. This domain bootstrapping strategy facilitates the transfer and broadening of meta-knowledge, transcending traditional single-perspective reasoning. 
Meta-reasoning enriches smaller models' reasoning abilities to dissect and interpret intricate dialogue scenarios, addressing the ``Value Selection Quandary'' and broadening their cognitive horizons. Moreover, this domain-agnostic feature aligns data distribution shifts across different domains, effectively reducing catastrophic forgetting.

Inspired by the powerful reasoning abilities of LLMs like LLaMA-2-70B~\cite{touvron2023llama} and ChatGPT \footnote{https://chat.openai.com/chat}, we introduce Reason-of-Select (RoS) distillation approach, designed to graft these advanced reasoning capabilities onto smaller models. We present a ``multi-value resolution'' strategy tailored to DST, prompting teacher LLMs to generate a ``selection chain'' that discerns and elaborates the best value choice from various options. Then, we distill this rationale into a smaller student model to boost its meta-reasoning capabilities, making it easier to deploy to terminal systems.

Moreover, to enhance faithful reasoning and reduce hallucinations in the teacher model, we integrate a schema-guided prompt, strengthening source-target correlation and semantic coherence. 
Our innovative Semantic Contrastive Reasoning Selection method uses semantic similarity to select the most accurate teacher-generated rationale, suitable even in black-box scenarios. Extensive experiments with various teacher and student model sizes demonstrate our method's exceptional Continual DST performance and robust cross-dataset generalization capabilities.

To summarize, our main contributions are:
\begin{itemize}[leftmargin=*,itemsep=2pt,topsep=0pt,parsep=0pt]
\item 
We propose a Reason-of-Select (RoS) distillation framework for Continual DST. Through enhancing domain-agnostic meta-reasoning capabilities, ambiguous value selection and catastrophic forgetting caused by data distribution shifts across domains can be effectively mitigated.

\item 
To alleviate the hallucination outputted by a teacher model in our annotation-free reasoning framework, we present a Semantic Contrastive Reasoning Selection strategy to obtain trust-worthy rationale from multiple candidates, ensuring faithful reasoning transfer.

\item 
Comprehensive experiments with two teacher models and four student models of varying sizes demonstrate superior performance and robust generalization of our proposed method.

\end{itemize}

%% file: 2.motivation.tex
Through an initial experiment with the same setups from prior work~\cite{zhu2022continual} (details in Section~\ref{sec:exp}), we employed two distinct backbone models, T5-small and LLaMA-7B, and observed a performance decline in Continual DST with turn ids increasing (Figure~\ref{fig:motivation} (a)). 
This excites two fundamental questions for obtaining better continual DST performance: 
(i) What crucial knowledge do current models lack for continual DST?
(ii) How can this missing knowledge be effectively integrated?

\subsection{Pinpointing the Value Selection Quandary}

\begin{figure}[t]
  \centering
  \subfigure[Avg. JGA score at each turn]{\includegraphics[width=0.59\linewidth]{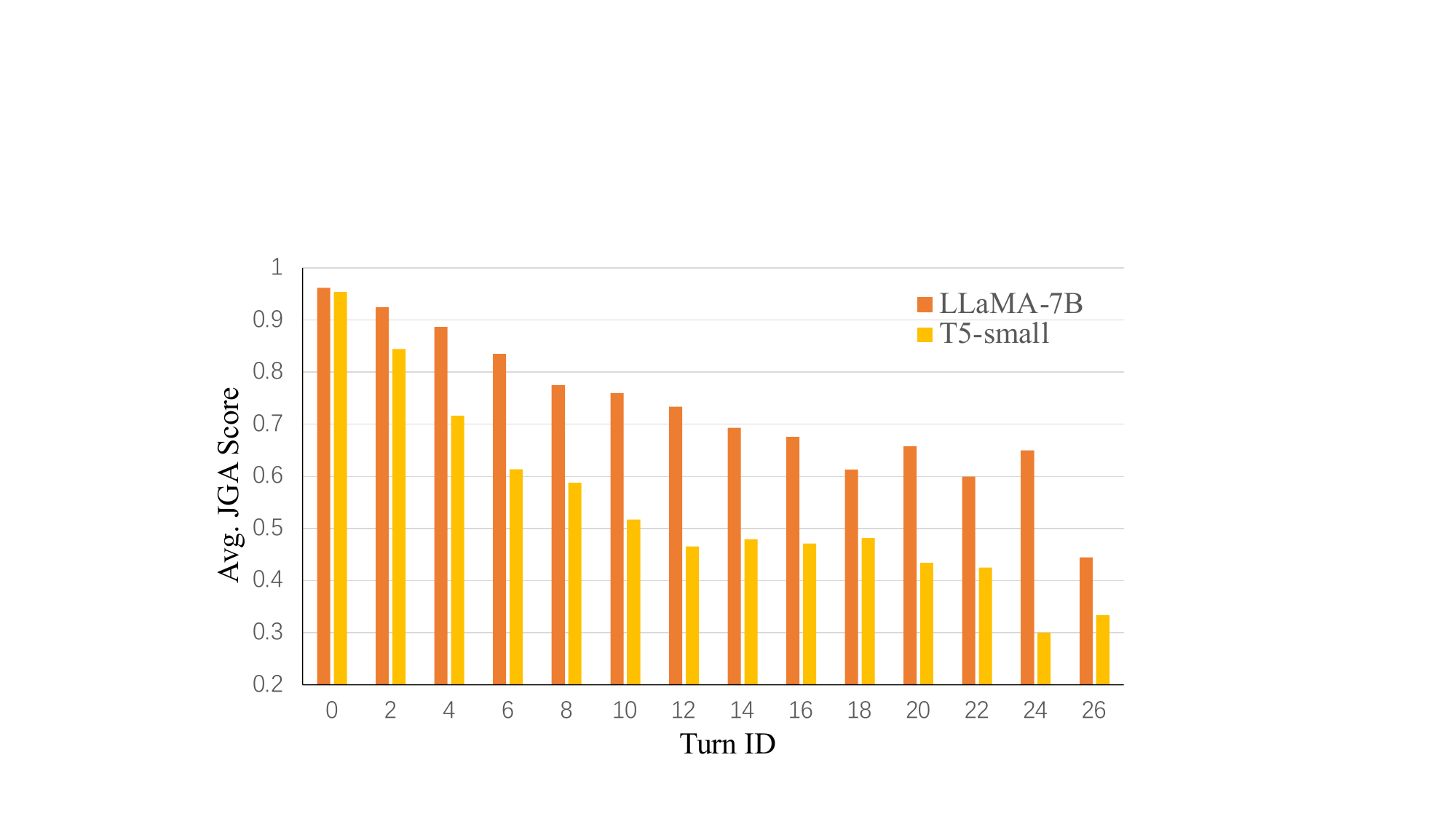}}
  \subfigure[Error rate for slots]{\includegraphics[width=0.39\linewidth]{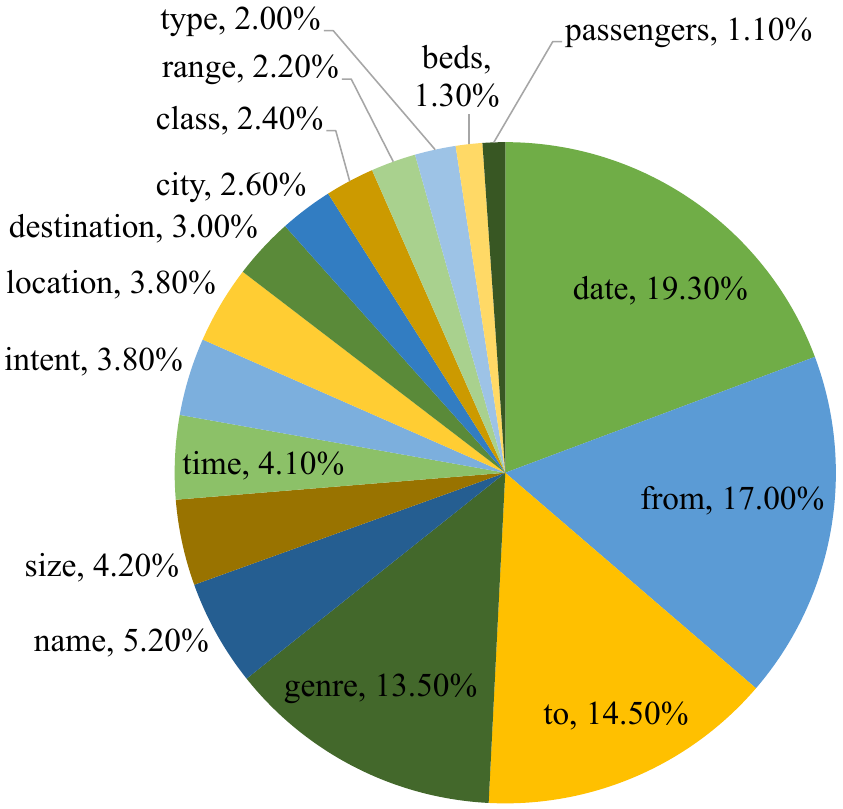}}
  \caption{Performance analysis of LLaMA-7B and T5-small in Continual DST task.
  }
  \label{fig:motivation}
\end{figure}

Upon analyzing the 874 incorrectly predicted slot samples in long dialogues (turn id > 10), we initially found that in 94.5\% of cases, the model usually returns values semantically relevant to the requested slot from the dialogue.
For example, a location-related slot often results in selecting any mentioned location.
Upon closer examination, we found that the model failed to select the correct value due to a significant capability loss, termed the ``Value Selection Quandary''. This issue occurs when the model struggles to reason out the correct value among multiple ambiguous or semantically similar values in a dialogue.

Examining 826 errors linked to this quandary, we observed a pattern: in 45\% of cases, models favored the most recently mentioned value (Figure \ref{fig:intro}). In 30\% of instances, they stuck with a previously chosen value, disregarding modifications in the dialogue. The remaining 25\% involved random selections from available values.
Moreover, Figure~\ref{fig:motivation}(b) classifies error rates across 49 slots, highlighting that slots regarding `date', `from', and `to' are most prone to errors. This is logical, as time or location slots in dialogues often present multiple values, leading to selection challenges.

These findings highlight a critical gap in models' ability to navigate the ``Value Selection Quandary'', presenting a significant avenue for enhancement.

\subsection{Bridging the Gap of Reasoning Ability}
Inspired by the robust reasoning capabilities exhibited by models exceeding 70B parameters \cite{li2023symbolic, wang-etal-2023-scott}, we evaluated their efficacy in DST task, specifically their ability to generate coherent rationales. As demonstrated in Table \ref{fig:teacher_reasoning}, ChatGPT's rationalizations for the <appointment\_date> slot confirm their proficiency in discerning and justifying value selections.

Our Reason-of-Select Distillation method, leveraging LLMs' selection logic, boosts smaller models' reasoning and decision-making abilities. This domain-agnostic approach effectively reduces forgetting in continual learning by teaching models to reason across diverse tasks and domains instead of merely imparting specific information or tasks.
In addition, we revealed the \emph{hallucination} in teacher models' rationales, such as references to non-existent dialogue elements (Table \ref{fig:teacher_reasoning}), which motivates a novel Semantic Contrastive Selection method (Section~\ref{sec:faithful}) ensuring the accuracy and relevance of the logic transferred to student models.

\begin{table}[t]
  \centering
  \includegraphics[width=1\linewidth]{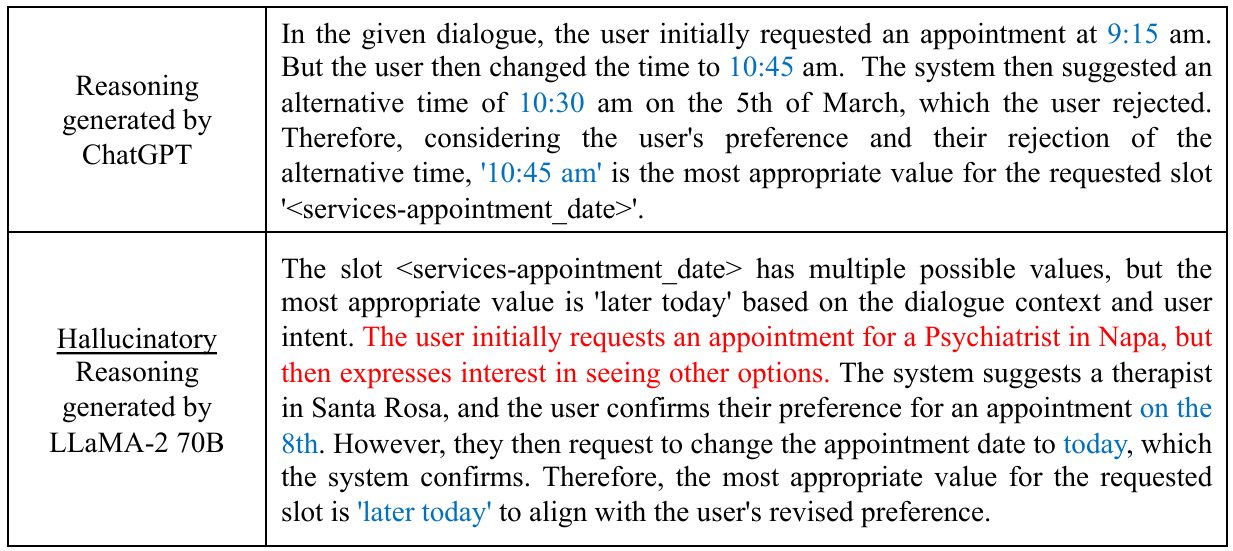}
  \caption{Examples of the reasoning processes generated by different teacher models, where blue font highlights possible values in the dialogue, and red font marks hallucinatory elements in the teachers' reasoning.}
  \label{fig:teacher_reasoning}
\end{table}

\begin{figure*}[t]
  \centering
  \includegraphics[width=0.9\linewidth]{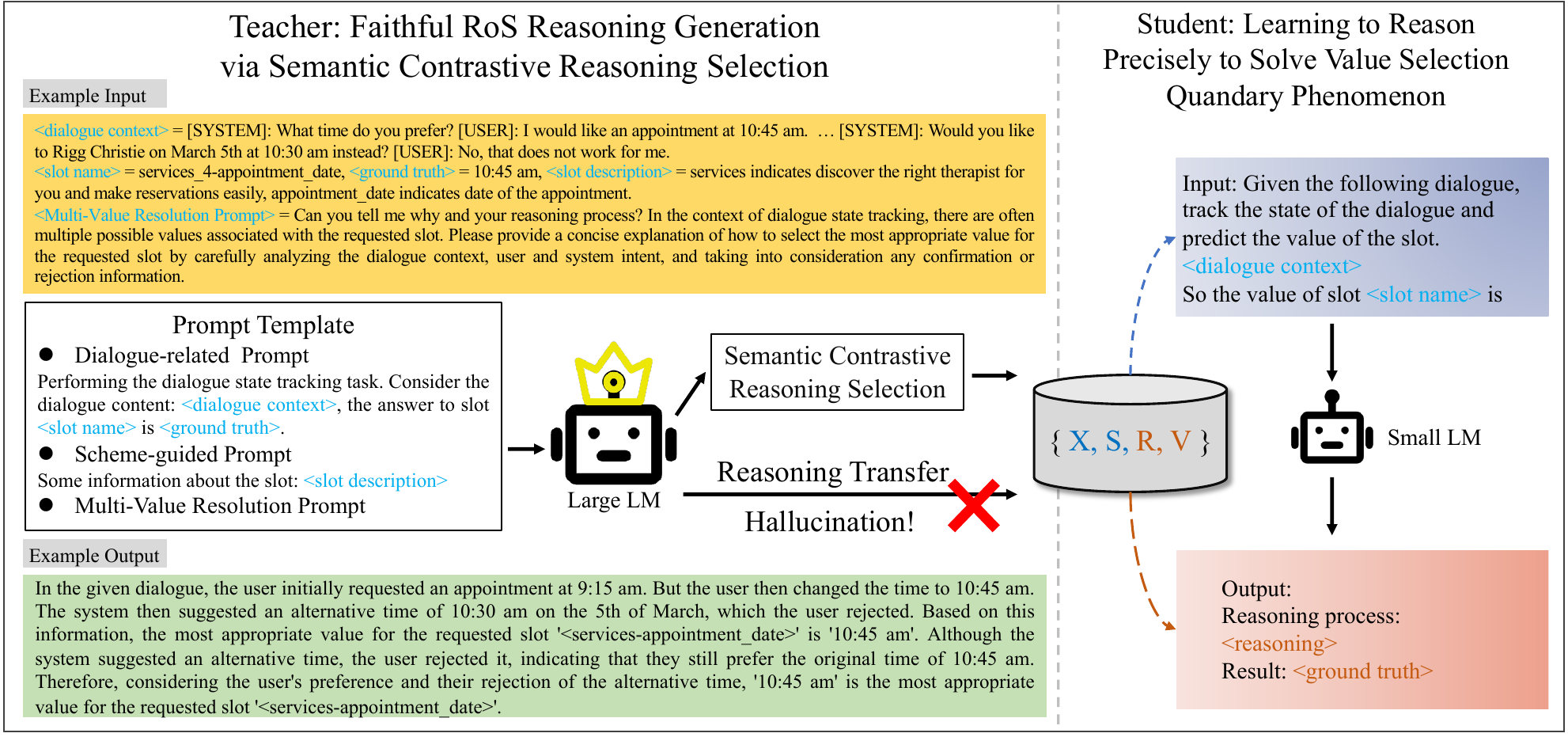}
  \caption{\textbf{Overview of the Reason-of-Select (RoS) Distillation method.} (a) Teacher: A large LM prompted to generate a faithful rationale given a dialogue context and the value for the request slot in the training set via the ``multi-value resolution'' strategy and Semantic Contrastive Reasoning Selection method. (b) Student: A small LM is fine-tuned to generate an accurate rationale and the corresponding value.
  }
  \label{fig:method}
\end{figure*}

%% file: 3.method.tex
\paragraph{Problem Formulation}
In continual DST, we train a model $f: \mathcal{X} \times \mathcal{T} \rightarrow \mathcal{Y}$ across a series of dialogue domains $\mathcal{T}_1,...,\mathcal{T}_K$. This model predicts the target $y$ based on input $x$ and task $\mathcal{T}_k \in \mathcal{T}$.
Within a specific task $\mathcal{T}_k$, a dialogue with $T$ turns of interactions between the system and the user can be represented as $\mathcal{X}_{T} = \left\{\left(A_{1}, U_{1}\right),\left(A_{2}, U_{2}\right) \ldots,\left(A_{T}, U_{T}\right)\right\}$, where $A$ represents the system response and $U$ represents the user input.
A predefined slot set\footnote{To specify the domain to which a slot belongs, a slot is defined as the concatenation of the specific domain and the slot name, e.g., ``<restaurant-area>''.} $\mathcal{S}=\left\{S_{1}, \ldots, S_{J}\right\}$ is provided, where $J$ is the total number of slots for task $\mathcal{T}_k$.
The objective of DST is to predict the dialogue state $\mathcal{B}_{t}$ based on the dialogue context $\mathcal{X}_{t}$. The dialogue state, $\mathcal{B}_{t}$, is represented as a set of (slot, value) pairs, denoted as $\mathcal{B}{t}=\left\{\left(S_{1}, V_{1}^{t}\right), \ldots,\left(S_{J}, V_{J}^{t}\right)\right\}$, where $V_{J}^{t}$ is the value associated with slot $S_{J}$ at turn $t$.
Essentially, the DST problem is defined as training a dialogue state tracker model $f: \mathcal{X}_{t} \oplus S_{j} \rightarrow V_{j}^{t}$, where $\oplus$ denotes simple text concatenation.

\subsection{Overview}

Our Reason-of-Select Distillation framework capitalizes on LLMs to create a faithful selection reasoning process through a `multi-value resolution' strategy and Semantic Contrastive Reasoning Selection. This enriched knowledge is then imparted to smaller student models for training, as illustrated in Figure \ref{fig:method}.


\subsection{Teacher's Reasoning Generation}

The process begins with a dialogue-centric prompt that includes dialogue content $\mathcal{X}$, the target slot $S_j$, and its value $V_j$ to derive a reasoning process $\mathcal{R}$.
While traditional prompts like ``Tell me why <$S_j$> is $V_j$'' or the famous ``Let’s think step by step'' prompt \cite{kojima2022large} generate reasonings that tend to merely highlight the location of the correct answer in the dialogue.

Our innovative ``multi-value resolution'' prompt $\mathcal{P}_R$ fosters a more elaborate reasoning process, enhancing the model's capability from mere location identification to an in-depth selection reasoning process, as illustrated in Figure \ref{fig:method}. This methodological advancement from basic identification to complex process reasoning represents a substantial leap in the model's cognitive capacities.

This stage's input-output representation is $\displaystyle f_{teacher}: \mathcal{X}_{t} \oplus S_{j} \oplus V_{j}^{t} \oplus \mathcal{P}_R \rightarrow \mathcal{R}_{j}^{t}$.
However, neural LMs often exhibit hallucinations — generating text with tenuous ties to the input \cite{ji2023survey, maynez2020faithfulness}, which our unsupervised annotation approach can exacerbate.
To counter this, we introduce the innovative Semantic Contrastive Reasoning Selection method, pivotal for accurately aligning reasoning with corresponding answers.

\subsection{Ensuring Faithful Teaching with Semantic Contrastive Reasoning Selection}\label{sec:faithful}
Our novel strategy employs a newly schema-guided prompt (yellow background in Figure \ref{fig:method}) to direct teacher models toward generating on-topic rationales, significantly improving the source-target correlation. 
Our Semantic Contrastive Reasoning Selection technique revolutionizes reasoning generation, creating a spectrum of candidates and choosing the most semantically aligned.
Inspired by contrastive decoding \cite{li2022contrastive} in text generation task, our method adapts this concept to suit black-box LLMs like ChatGPT.
We introduce strategic input perturbations to simulate and correct reasoning errors, refining the teacher model's output.

\begin{figure}[t]
  \centering
  \includegraphics[width=0.9\linewidth]{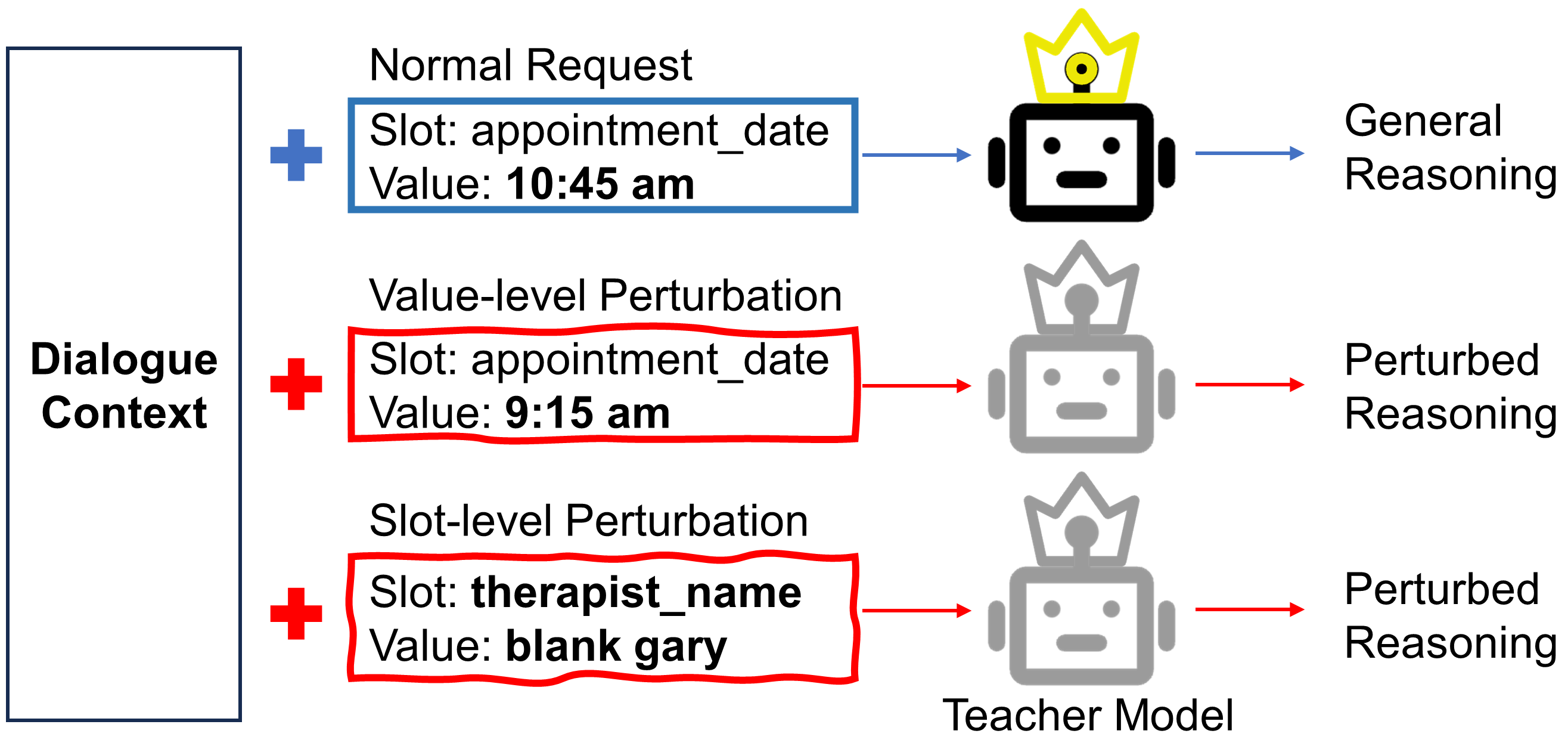}
  \caption{Demonstration of value-level and slot-level perturbations to elicit diverse negative reasonings.}
  \label{fig:perturbation}
\end{figure}

\begin{figure}[t]
  \centering
  \includegraphics[width=0.9\linewidth]{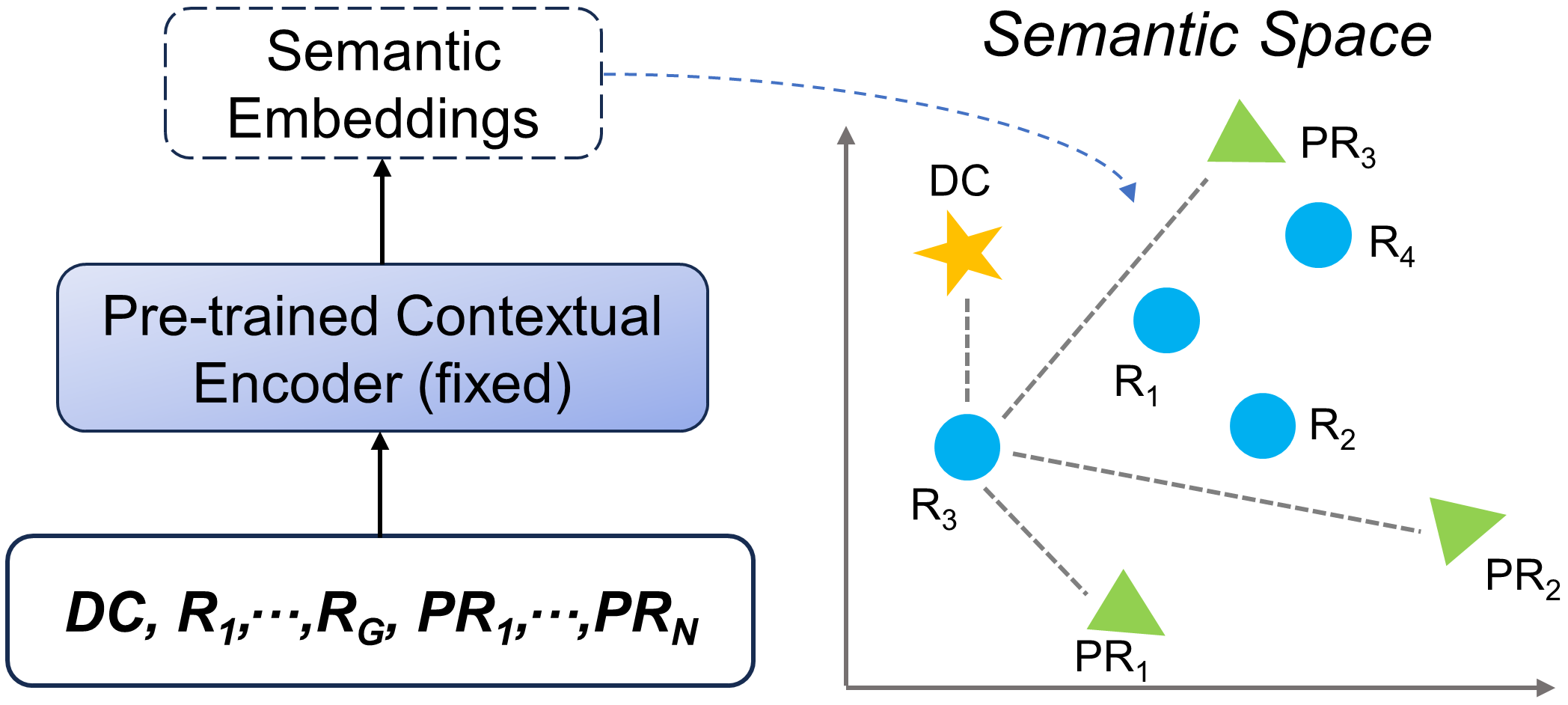}
  \caption{Illustration of the Semantic Contrastive Reasoning Selection method.}
  \label{fig:contrastive}
\end{figure}

The procedure commences with the LLM generating $G$ diverse reasonings ($\mathcal{R}_1, \ldots, \mathcal{R}_G$). 
We then introduce two forms of perturbation: value-level and slot-level, as illustrated in Figure \ref{fig:perturbation}. While value-level perturbation subtly modifies the ground truth value, slot-level perturbation involves completely replacing the slot-value pair. This latter proves particularly effective in inducing a range of reasoning responses, from logical to nonsensical, thereby enriching our negative sample pool with $N$ perturbated reasonings ($\mathcal{PR}_1, \ldots, \mathcal{PR}_N$). Positive samples are represented by the dialogue-centric prompt, denoted as $\mathcal{DC}$.
Utilizing a fixed, pre-trained contextual encoder, Sentence-transformers \cite{reimers2019sentence}, we then transform these $\mathcal{R}$, $\mathcal{PR}$ and $\mathcal{DC}$ into semantic representations within space $\mathcal{E}$, as illustrated in Figure \ref{fig:contrastive}.
 
The crux of our method lies in selecting the most appropriate reasoning from the generated $G$ candidates, a decision based on their semantic proximity to the positive samples and divergence from the negative samples. 
This selection process is formulated as an elegant optimization problem:

\begin{equation}
\begin{aligned}
& \min Distance(\mathcal{R}_i, \mathcal{DC}) \\
& \textrm{s.t.} \max \sum_{n=1}^{N}{Distance(\mathcal{R}_i, \mathcal{PR}_n)} \\
\end{aligned}
\end{equation}

To quantify the relationship between each reasoning and the sample sets, we introduce a sophisticated scoring mechanism:

\begin{equation}
Score(\mathcal{R}_i)=\frac{\exp \left(d\left(\mathcal{R}_i, \mathcal{DC}\right) / \tau\right)}{\sum_{n=1}^{N} \exp \left(d\left(\mathcal{R}_i, \mathcal{PR}_n\right) / \tau\right)}
\end{equation}
where $d$ is a distance function and $\tau$ is a temperature scalar.
By computing a score for each reasoning, we discern the optimal choice with the lowest score for training the student model based on its alignment with actual dialog context and its divergence from perturbated content.

\subsection{Training Student Models via Reasoning-Enhanced Data}

Armed with the annotated dataset $ \left\{\mathcal{X}, S, \mathcal{R}, V\right\}$, we proceed to train a smaller student model within the self-rationalization framework, emphasizing both predictive and explanatory skills.
This approach departs from prior post-rationalization models, where rationales are formulated post-prediction or those employing a multi-task format, treating rationale creation as an auxiliary task.


The student model is conditioned to generate a sequence that merges rationale tokens with the corresponding answer tokens in response to a given dialogue context and slot request, as demonstrated in the right part in Figure \ref{fig:method}. This task is executed by fine-tuning a pre-trained language model, with ``silver'' data derived from the teacher model. 
We employ a standard language modeling loss termed factual reasoning loss:
\begin{equation}
\mathcal{L}_{factual}=-\sum_{j}^{J} \log P\left(\mathcal{R}_{j}, V_{j} \mid \mathcal{X}, S_{j} \right)
\end{equation}

%% file: 4.exp.tex
\subsection{Experimental Setup}
\paragraph{Dataset}
Our experiment employs the Schema-Guided Dialog dataset (SGD)~\cite{rastogi2020towards}, encompassing 44 services across 19 domains with slot descriptions. Adhering to the Continual DST setup by \citet{zhu2022continual}, we focus on single-service dialogs, randomly selecting 15 tasks from 44. Each service varies in training sample size and slots. To ensure robustness, we experiment with five task orders via random permutations, aligning with prior studies. Appendix \ref{sec:dataset} details the data statistics, task selection, and orderings.

\paragraph{Evaluation Protocol}
We assess the DST performance using the widely adopted Joint Goal Accuracy (JGA) metric \cite{wu-etal-2019-transferable}, which demands accurate predictions for all slot values. 
We denote $a_{j,i}$ as the JGA on the test set of task $\mathcal{T}_i$ right after training on task $\mathcal{T}_j$.
CL performance is assessed using three metrics from \citet{zhu2022continual}: \textit{(i)} $\displaystyle \mathbf{Avg. JGA} =\frac{1}{K} \sum_{i=1}^{K} a_{K, i} $, representing the average JGA across all tasks after training on the final task $\mathcal{T}_K$.
\textit{(ii)} Forward Transfer ($\mathbf{FWT}$) $\displaystyle = \frac{1}{K-1} \sum\limits_{i=2}^{K} a_{i-1, i}$, evaluating generalization by measuring zero-shot performance, and \textit{(iii)} Backward Transfer ($\mathbf{BWT}$) $\displaystyle = \frac{1}{K-1} \sum\limits_{i=1}^{K-1} a_{K, i}-a_{i, i}$, quantifying resistance to forgetting by assessing the influence of new learning on previous tasks.

\newcommand{\tabincell}[2]{\begin{tabular}{@{}#1@{}}#2\end{tabular}}
\begin{table*}[t]
\centering
\scalebox{0.8}{
\begin{tabular}{l|cl|ccc|ccc}
\toprule
Method&  Teacher  & Student & Avg. JGA&  FWT&  BWT&  +Memory&  +Params& +Reg.\\
\midrule

\rule{0pt}{4pt}\emph{Fine-tune} & -& \multirow{7}*{\tabincell{c}{\emph{T5-small}\\}} & $44.1_{0.9}$ & $8.3_{1.0}$ & $-36.6_{3.9}$ & - & - & - \\
\rule{0pt}{8pt}\emph{EWC}  &  -&  & $47.9_{1.1}$ & $8.4_{0.9}$ & $-38.1_{4.1}$ & \checkmark & \checkmark & \checkmark \\
\rule{0pt}{8pt}\emph{Replay}  &  -&  & $58.6_{3.5}$ & $10.9_{0.5}$ & $-3.2_{2.3}$ & \checkmark & - & - \\
\rule{0pt}{8pt}\emph{AdapterCL}   &  -&  & $49.8_{1.7}$ & - & - & - & \checkmark & - \\
\rule{0pt}{8pt}\emph{CPT} &  -&  & $61.2_{2.5}$ & $13.7_{0.8}$ & $\textbf{0.5}_{0.4}$ & \checkmark & \checkmark & \checkmark \\
\rule{0pt}{8pt}\emph{DST-EGQA} &  -&  & $55.5_{3.5}$ & $23.6_{2.1}$ & $-19.1_{4.2}$ & - & - & - \\
\rule{0pt}{8pt}\emph{\quad+ Dialogue Memory} & -&  &  $68.9_{0.3}$ & $22.5_{1.8}$ & $-5.9_{1.9}$ & \checkmark & - & - \\
\midrule


\rule{0pt}{8pt} \emph{RoS (ours)}  &  \multirow{2}*{\tabincell{c}{\emph{LLaMA-2-70B}\\}} & \multirow{2}*{\tabincell{c}{\emph{T5-small}\\}} & $59.0_{3.9}$ & $25.5_{2.0}$ & $-17.9_{3.7}$ & - & - & - \\
\rule{0pt}{8pt} \emph{\quad+ Dialogue Memory} &   &  & $\textbf{72.1}_{0.8}$ & $26.7_{2.0}$ & $-2.6_{1.5}$ & \checkmark & - & - \\
\rule{0pt}{8pt} \emph{RoS (ours)}  &  \multirow{2}*{\tabincell{c}{\emph{ChatGPT}\\}} & \multirow{2}*{\tabincell{c}{\emph{LLaMA-7B}\\}} & $68.7_{4.1}$ & $\textbf{51.9}_{1.7}$ & $-8.5_{3.8}$ & - & - & - \\
\rule{0pt}{8pt} \emph{\quad+ Dialogue Memory} &   &  & $\textbf{74.2}_{3.7}$ & $\textbf{52.7}_{1.5}$ & $\textbf{-2.4}_{2.7}$ & \checkmark & - & - \\

\midrule
\rule{0pt}{8pt}\emph{CPT Multi-task} & -& \emph{T5-base} &  $64.0_{1.9}$ & - & - & - & \checkmark & \checkmark \\
\rule{0pt}{8pt}\emph{DST-EGQA Multi-task} &  -& \emph{T5-small} & $74.2_{1.8}$ & - & - & - & - & - \\
\rule{0pt}{8pt}\multirow{2}*{\tabincell{c}{\emph{RoS Multi-task}\\}} 
&  \emph{LLaMA-2-70B} & \emph{T5-small} & $76.3_{0.5}$ & - & - & - & - & - \\
\rule{0pt}{8pt} &   \emph{ChatGPT} & \emph{LLaMA-7B}  & $78.9_{0.3}$ & - & - & - & - & - \\
\bottomrule
\end{tabular}}
\caption{CL metric results and reliance on other continual learning techniques. Means and standard variances are reported. We compare models sequentially trained on 15 tasks from the SGD dataset and aggregate results across five domain permutations. The last four rows provide the multi-tasking results, which serve as an upper bound. All rows that use memory are with $M$ = 50.}
\label{tbl:result}
\end{table*}

\paragraph{Baselines}
We evaluate our model against existing Continual DST baselines:
\textbf{\emph{Fine-tuning:}} Continuously fine-tune the model on new task data.
\textbf{\emph{Replay:}}
Stores $\left| M \right|$ instances per task $\mathcal{T}_i$ in memory $M_i$ for joint training with new tasks.
\textbf{\emph{EWC \cite{kirkpatrick2017overcoming}:}}
Maintain a memory but leverage it to compute the Fisher information matrix for regularization.
\textbf{\emph{AdapterCL \cite{houlsby2019parameter}:}}
Freeze the pre-trained model and independently train a residual Adapter for each task.
\textbf{\emph{Continual Prompt Tuning (CPT) \cite{zhu2022continual}:}}
Freeze the backbone model and continually train soft prompts with knowledge transfer in both forward and backward directions.
\textbf{\emph{DST-EGQA \cite{cho2023continual}:}}
Reformulate DST as a Question-Answering task using retrieval-augmented in-context learning.

We also include our method with memory replay and multi-task learning as performance caps.

\paragraph{Training Details}
ChatGPT (using the gpt-3.5-turbo API) and LLaMA-2-70B serve as teacher models for fine-tuning smaller student models including T5-small, T5-base, FlanT5-XL, and LLaMA-7B, utilizing generated rationales. The teacher model's temperature is set to 0.7, generating five candidate reasonings ($G$ = 5) alongside three value-level and three slot-level perturbed negative reasonings ($N$ = 6).
In Eq.2, the parameter $\tau$ is set to 0.8, employing the Euclidean distance metric.
The memory size per task $\left| M \right|$ is maintained at 50, in line with previous studies \cite{zhu2022continual}. Detailed training specifications are provided in Appendix \ref{sec:imp}.

\subsection{Main Results}
Table \ref{tbl:result} presents the results from various methodologies. Key findings include:

\paragraph{Injecting Reasoning Knowledge Effectively Enhances CL Performance}
Our RoS method significantly surpasses standard fine-tuning on T5-small, raising the Avg.JGA from 44.1\% to 59.0\%, and showing gains in both FWT and BWT. This highlights the value of integrating reasoning into Continual DST.
Compared with the erstwhile top-performing DST-EGQA, we achieve a new SOTA performance in all metrics, notably increasing Avg. JGA from 55.5\% to 59.0\% without extra memory, demonstrating our approach's effectiveness in mitigating historical task forgetting.
Furthermore, when memory is available, RoS jumps from 59.0\% to 72.1\%, even outstripping the CPT's multi-task performance. Among various student models, RoS with LLaMA-7B, fine-tuned with ChatGPT's reasoning, stands out, substantially lifting the Avg. JGA from 59.0\% to 68.7\%, with a remarkable FWT improvement from 25.5\% to 51.9\%.

\begin{figure}[t]
  \centering
  \includegraphics[width=0.8\linewidth]{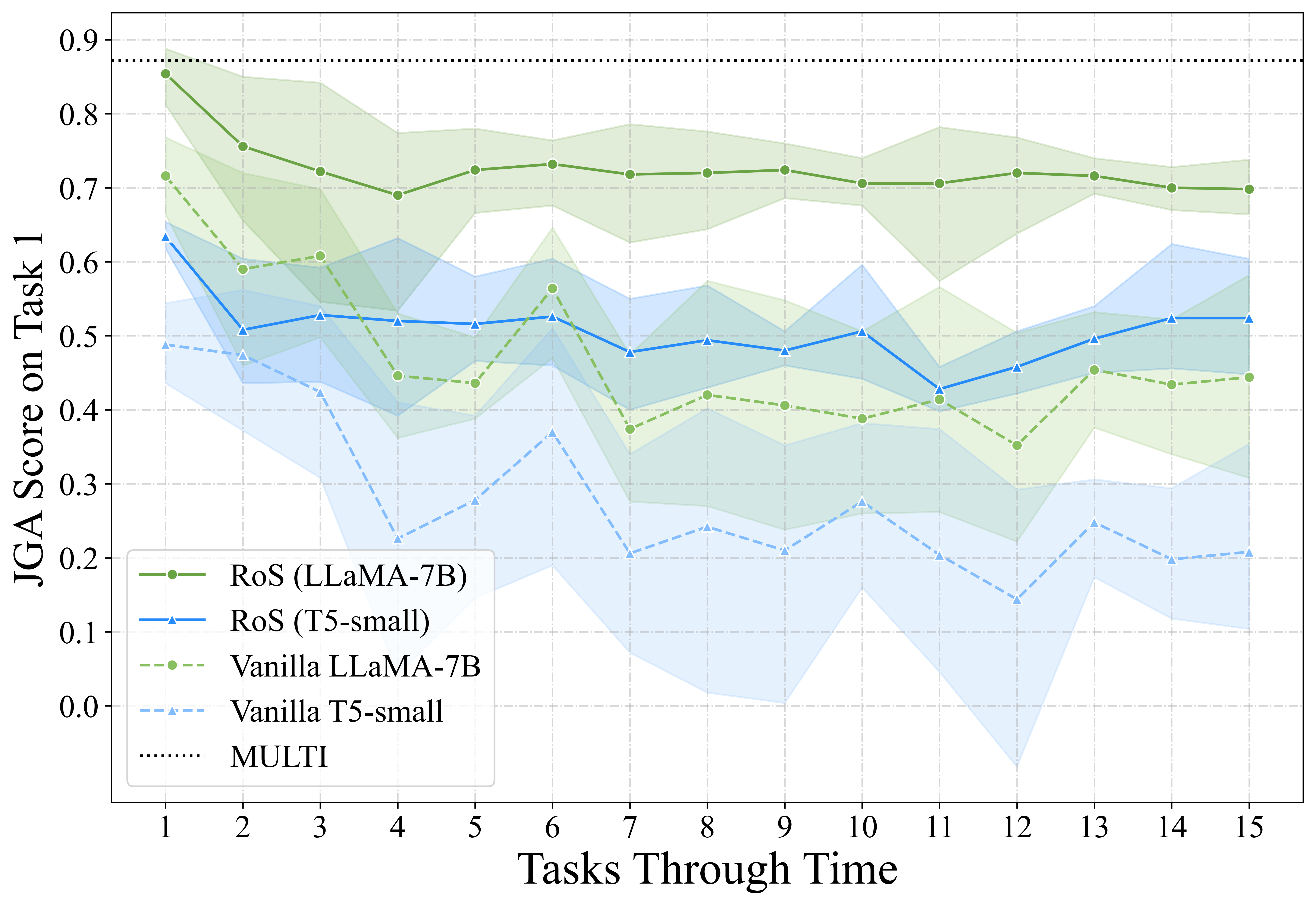}
  \caption{Task 1 performance trajectory during continual DST learning process.}
  \label{fig:forgetting}
\end{figure}

Figure \ref{fig:forgetting} shows the models' ability to mitigate catastrophic forgetting by assessing their performance on the initial task post successive task learnings. Models with RoS reasoning skills exhibit a slower forgetting rate, with a 15\% average drop in performance in both T5-small and LLaMA-7B. In contrast, vanilla backbone models show a steeper decline, averaging a 28\% performance dip, highlighting the importance of domain-agnostic reasoning skills in sustaining historical task performance.


\paragraph{RoS Distillation Exhibits Strong Generalization Ability}
Our method significantly improves the FWT metric, indicative of robust generalization and zero-shot learning capabilities. It bridges the distribution gap across domains by creating a systematic reasoning chain, evident in handling semantically similar but differently described slots, like <services-appointment\_date> and <hotels-check\_in\_date>. Unlike traditional models, ours utilizes reasoning to recognize similarities, enhancing adaptability to unseen slots.

\begin{table}[t]
\centering
\scalebox{0.65}{
\begin{tabular}{llcccc}
\toprule
Method & Backbone & Attraction &  Restaurant  &Train & Average \\
\midrule
\rule{0pt}{8pt}\emph{CPT} & \emph{T5-base} &10.05 & 19.37 &3.34 &10.92\\
\midrule
\rule{0pt}{4pt}\multirow{2}*{\tabincell{c}{\emph{Fine-tune}\\}}   & \emph{T5-small}&8.40 &12.87 &2.94  &8.07  \\

\rule{0pt}{4pt}  & \emph{LLaMA-7B}  &40.47 &52.73 &18.75  &37.32  \\
\midrule
\rule{0pt}{8pt}\multirow{2}*{\tabincell{c}{\emph{RoS}\\}} & \emph{T5-small} &\textbf{15.74} &\textbf{24.07}  &\textbf{4.05} &\textbf{14.62}\\

\rule{0pt}{8pt}  & \emph{LLaMA-7B} &\textbf{42.84} &\textbf{59.94}  &\textbf{25.57} &\textbf{42.78}\\
\bottomrule
\end{tabular}}
\caption{Zero-Shot performance on the 16th cross-dataset task using the MultiWOZ 2.4 dataset.}
\label{tbl:crossdataset}
\end{table}

To further evaluate this generalization, we introduce a new 16th task using MultiWOZ 2.4 data \cite{ye-etal-2022-multiwoz}.
The zero-shot performance on this task follows the sequential training of the first 15 tasks from the SGD, as shown in Table \ref{tbl:crossdataset}.

Compared to similar parameter-sized backbones, RoS (T5-small) outperforms the T5-base CPT baseline, increasing average JGA by 3.7\%, from 10.92\% to 14.62\% across domains. This is particularly notable in the Attraction domain, with JGA rising from 10.05\% to 15.74\%. Compared to various vanilla backbones, the improvement due to domain-agnostic reasoning averages a 6\% increase, further confirming RoS's robust generalization.

To assess the hallucination rate of the teacher model, we utilized SelfCheckGPT~\cite{manakul2023selfcheckgpt} and employed BERTScore for evaluation. BERTScore provides scores ranging from [0.0, 1.0], where higher values indicate a greater likelihood of non-factual content, implying an increased probability of hallucination.
Our evaluation included the average BERTScore for the initial generation of five candidate reasonings by the teacher models, the BERTScore after applying the Semantic Contrastive Reasoning Selection method, and the hallucination rate of the distilled student. The summarized results are presented in the Table \ref{tbl:hallucination}.

\begin{table}[t]
\centering
\scalebox{0.75}{
\begin{tabular}{lcc}
\toprule
\multirow{2}*{\tabincell{c}{Teacher Model\\}} & \multirow{2}*{\tabincell{c}{Initial \\ BERTScore (\%)}}& \multirow{2}*{\tabincell{c}{BERTScore After\\ Contrastive Selection (\%)}}\\
&& \\

\midrule
\rule{0pt}{4pt}ChatGPT  & 	19.3 &14.2\\
\rule{0pt}{8pt}LLaMA-2-70B & 21.7  &	15.9 \\

\bottomrule
\end{tabular}}
\caption{Hallucination rate of the reasoning generated by teacher models after our contrastive selection method.}
\label{tbl:hallucination}
\end{table}

Table \ref{tbl:hallucination} illustrates that for both teacher models, ChatGPT and LLaMA-2-70B, the average BERTScore for the initial five reasoning candidates were 19.3\% and 21.7\%, respectively. After applying the Semantic Contrastive Reasoning Selection method, the BERTScore decreased to 14.2\% and 15.9\%, affirming the effectiveness of our proposed approach in enhancing the factual of reasoning.
Regarding the student models, there is no significant difference in BERTScore when distilled with different teacher models. Additionally, we observed that the quality of reasoning generated by the student models improves as the model size increases.

\subsection{Ablation Study}
\begin{figure}[t]
  \centering
  \includegraphics[width=1.0\linewidth]{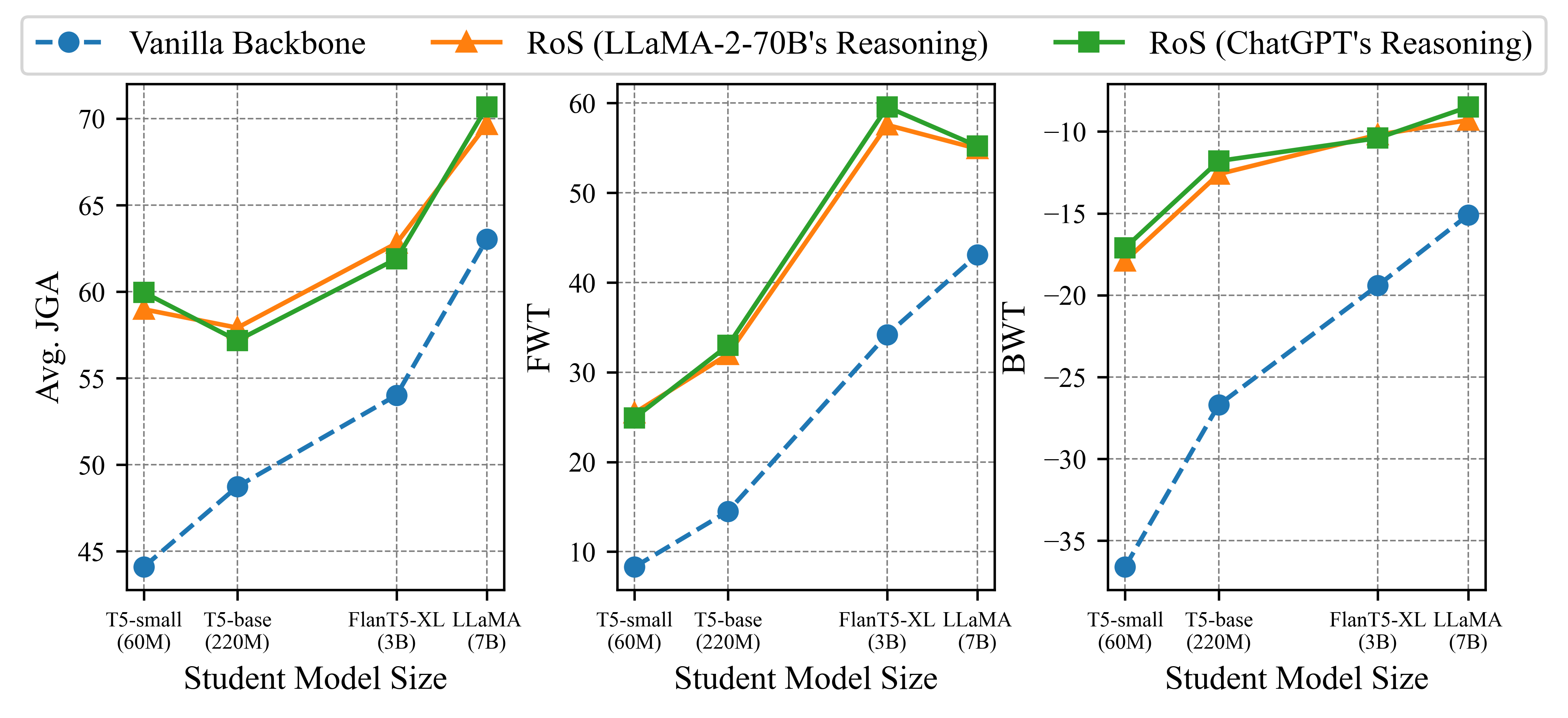}
  \caption{Comparative task performance across varying sizes of teacher and student models.}
  \label{fig:different_size}
\end{figure}

\paragraph{The Role of Model Size Variation in Teacher and Student Models}
Our study investigates the effect of teacher and student model sizes on performance. Figure \ref{fig:different_size} shows a clear trend: larger student models, like LLaMA-7B, outperform smaller ones, such as T5-small. 
We first evaluate LLaMA-7B's continual learning in DST, finding it outperforms most T5-small or T5-base-based methods.
However, applying our RoS method with reasoning is the most notable enhancement, especially in smaller models.
For example, integrating reasoning into T5-small yields an impressive 15\% average increase in all evaluation metrics.
Contrary to our initial assumption that students guided by ChatGPT might surpass those instructed by LLaMA-2-70B, the results indicate equal reasoning quality from both teacher models, suggesting their similar effectiveness in enhancing student model learning.

\paragraph{Effectiveness of Semantic Contrastive Reasoning Selection Method}

To assess our method's ability to reduce reasoning hallucinations, we compared three reasoning selection strategies: random selection from $G$ candidates and selecting the highest and lowest scoring (ours) reasonings, as per Equation 2. The results are detailed in Table \ref{tbl:ablation_hallucination}.

Initial results show that all methods significantly outperform the vanilla LLaMA, with even random selection improving Avg. JGA by 4.94\%. This highlights the inherent reasoning quality in current teacher models.
Yet, refining the selection to eliminate hallucinatory reasonings and aggregate higher-quality reasonings further enhanced performance. Compared to random selection, our method shows additional gains of 3.2\%, 0.9\%, and 2.8\% in the evaluated metrics.
In contrast, selecting the highest-scoring result, i.e., favoring less effective reasoning, resulted in reduced performance.
These findings robustly affirm the effectiveness of our semantic similarity-based selection approach.

\begin{table}
\centering
\scalebox{0.75}{
\begin{tabular}{lccc}
\toprule
Models & Avg. JGA & FWT & BWT\\
\midrule
\rule{0pt}{6pt} \emph{LLaMA-7B (backbone)} &63.04 & 43.10 & -15.10 \\
\rule{0pt}{8pt}\hspace{0.5cm} \emph{+ Max Score} & 65.72  & 49.05  & -13.34 \\
\rule{0pt}{8pt}\hspace{0.5cm} \emph{+ Random Selection} & 66.98 & 51.14 & -11.73 \\
\rule{0pt}{8pt} \hspace{0.5cm}\textbf{\tabincell{c}{\emph{+ Min Score (Ours)}}} & \textbf{70.17} & \textbf{\tabincell{c}{52.06}}  & \textbf{\tabincell{c}{-8.90}}\\
\bottomrule
\end{tabular}}
\caption{Ablation study on the effectiveness of strategies to alleviate hallucination in teacher model's reasoning.}
\label{tbl:ablation_hallucination}
\end{table}

\subsection{Case Study}

\begin{table}[t]
  \centering
  \includegraphics[width=1\linewidth]{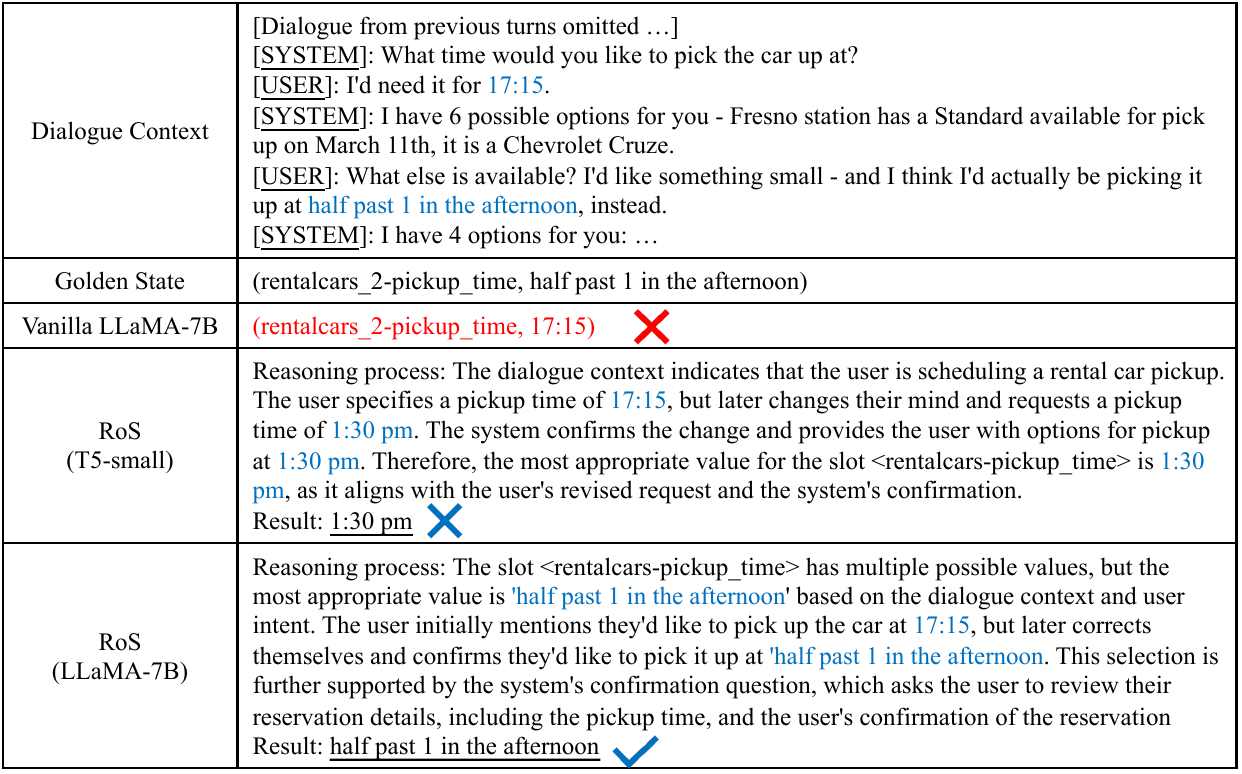}
  \caption{Exemplary dialogue from the SGD test set with two possible values, with predictions by the vanilla LLaMA-7B and our meta-reasoning augmented method applied to two distinct backbones.
  }
  \label{fig:casestudy}
\end{table}

Table \ref{fig:casestudy} presents a key test case and various methods' predictions to showcase student models' advanced reasoning capabilities. 
The dialogue involves two values for the <pickup\_time> slot: ``17:15'' and a later update to ``half past 1 in the afternoon''. The vanilla LLaMA model incorrectly selects ``17:15'', while our model effectively identifies all possible values and accurately determines the correct one.
Interestingly, our RoS T5-small model's output of ``1:30 pm'' mirrors the ground truth but is marked incorrect due to current evaluation criteria. This highlights a challenge in matching model outputs with expected formats, suggesting a direction for future refinement.




%% file: 5.related.tex
\subsection{Knowledge Distillation from LLMs}
Knowledge distillation in NLP, crucial for transferring insights from larger teacher models to smaller student models, has evolved recently, especially in extracting reasoning from LLMs.
Recent studies \cite{hsieh2023distilling, wang2022self, yang2023kerprint, xu2023vecocare} highlight LLMs' role in bolstering smaller models. \citet{li2023symbolic} furthered this with a symbolic Chain of Thought (CoT) distillation, enhancing smaller models through CoT prompting.
Despite these advancements, the faithfulness of generated rationales, critical for student models' behavior, often must be addressed. 
\citet{wang-etal-2023-scott} proposed a contrastive decoding method to reduce hallucinations, but its reliance on teacher model logits limits its applicability for black-box LLMs. Current reasoning distillation methods are unsuitable for DST task and lead to unproductive reasoning processes.

To overcome these issues, our method focuses on the Value Selection Quandary in DST, introducing a `multi-value resolution' prompt and a Semantic Contrastive Reasoning Selection method, enhancing knowledge transfer accuracy and relevance.

\subsection{Continual Dialogue State Tracking}
Continual Learning in task-oriented dialogue systems, focusing on mitigating catastrophic forgetting, has employed various methods such as architecture-based \cite{shen2019progressive, geng2021continual, xu2023seqcare}, rehearsal-based \cite{rebuffi2017icarl, hou2019learning, lu2021getting}, and regularization-based \cite{li2017learning, feng2024tasl, lu2021engage}. In DST, contributions like \citet{madotto2020continual} and \citet{liu2021domain} have utilized these CL strategies, with \citet{zhu2022continual} introducing Continual Prompt Tuning (CPT) to fine-tune domain-specific soft prompts.
Furthermore, DST-EGQA \cite{cho2023continual} employs a question-answering framework based on example-guided learning. However, its reliance on fixed QA templates may limit adaptability across diverse domains. Our method, in contrast, offers enhanced time efficiency and flexibility during testing, eliminating the need for sample retrieval, thus presenting a more efficient solution.

%% file: 6.conclusion.tex

This study introduces the Reason-of-Select (RoS) distillation method to address the Continual DST task's catastrophic forgetting and ``Value Selection Quandary'' challenges. Enhancing smaller models with a novel domain-agnostic `meta-reasoning' capability effectively broadens their reasoning horizon and aligns data distribution shifts across different domains.
Two innovative methods, the ``multi-value resolution'' strategy and Semantic Contrastive Reasoning Selection, further strengthen RoS by ensuring reliable knowledge transfer. Extensive experiments demonstrate our method's superior performance and robust generalization.

%% file: 7.limitations.tex

In contrast to the conventional knowledge distillation process, our approach necessitates additional computational resources during the preparation of training data and the training phase of the student model. Firstly, our Contrastive Reasoning Selection method requires the generation of multiple candidate reasonings, along with several value-level and slot-level perturbation reasonings. This process is more time-consuming than standard rational distillation, as it involves extensive data generation by the teacher model. Secondly, training student models with rationales introduces a minor increase in computational overhead during the training period. However, it’s important to note that these expenses are one-time costs. When it comes to testing, our model, already adept at reasoning, does not incur extra time for predictions, and the time involved in generating reasoning in the test phase is almost inconsequential.

%% file: 8.appendix.tex
\section{Description of Prompt Templates}
\label{sec:appendix}
\subsection{Prompt Templates for Generating Teacher Model Reasonings}
Below, we present a specific example of the prompt used to elicit reasoning generation from the teacher model.

\noindent 
\{

\noindent 
\texttt{\textbf{``instruction'':} Just return a concise reasoning process.}

\noindent 
\texttt{\textbf{``input'':} Performing the dialogue state tracking task. Consider the dialogue content: ``[Previous dialogue omitted ...] [USER]: I would like an appointment at 10:45 am. [SYSTEM]: When is the appointment for? [USER]: The appointment is for the 11th of March. [SYSTEM]: Booking appointment with Rigg Christie on March 11th at 10:45 am. [Remaining dialogue omitted ...]'', the answer to slot <services-appointment\_date> is `10:45 am’. }

\texttt{Some information about the slot: services indicates Discover the right therapist for you and make reservations easily, appointment\_date indicates Date of the appointment.}

\texttt{Can you tell me why and your reasoning process? In the context of dialogue state tracking, there are often multiple possible values associated with the requested slot. Please provide a concise explanation of how to select the most appropriate value for the requested slot by carefully analyzing the dialogue context, user and system intent, and taking into consideration any confirmation or rejection information.}

\noindent 
\}\vspace{12pt}

\subsection{Prompt Templates for Fine-Tuning the Student Model}
Below, we provide a specific example of the prompt used for fine-tuning the student model.

\noindent 
\{

\noindent 
\texttt{\textbf{``instruction'':} Given the following dialogue, track the state of the dialogue and predict the value of the slot <alarm\_1-new\_alarm\_name>.}

\noindent 
\texttt{\textbf{``input'':} [USER]: I want to check the alarms I have. [SYSTEM]: There are 2 alarms which you have set currently, with one of the alarms being at 6:30 am, and it is called Wake Up. [USER]: Alright, that is good. [SYSTEM]: Are you interested to add another alarm? [USER]: Actually I am, I do want to add another alarm. I want the alarm to be called Grocery run. }

\texttt{[slot] alarm\_1-new\_alarm\_name, it indicates Name to use for the new alarm. So the value of slot <alarm\_1-new\_alarm\_name> is}

\noindent 
\texttt{\textbf{``output'':} grocery run}

\noindent 
\}\vspace{12pt}


\section{Dataset Statistics}
\label{sec:dataset}
Here, we offer a detailed description of the dataset used in Continual DST~\cite{feng2022spatial}. Table \ref{tab:dataset_count} displays the number of slots for each of the 15 services used in our experiments and the count of samples in the training, validation, and test sets. Table \ref{tab:dataset_order} illustrates the training sequence for these 15 tasks in the context of continual learning.

\begin{table*}[t]
\centering
\begin{tabular}{ccrrrrrrrrr}
\toprule
Task ID &
  Service &
  \multicolumn{1}{c}{\# Slots} &
  \multicolumn{3}{c}{\# Dialogs} &
  \multicolumn{3}{c}{\# Samples} &
  \multicolumn{2}{c}{Avg. tokens} \\ \midrule
\textit{} &
  \textit{} &
  \multicolumn{1}{c|}{\textit{}} &
  \multicolumn{1}{c}{\textit{Train}} &
  \multicolumn{1}{c}{\textit{Dev}} &
  \multicolumn{1}{c|}{\textit{Test}} &
  \multicolumn{1}{c}{\textit{Train}} &
  \multicolumn{1}{c}{\textit{Dev}} &
  \multicolumn{1}{c|}{\textit{Test}} &
  \multicolumn{1}{c}{\textit{Context}} &
  \multicolumn{1}{c}{\textit{Query}} \\ \midrule
30 & services\_4    & \multicolumn{1}{c|}{5}  & 86  & 13 & \multicolumn{1}{r|}{25}  & 680  & 97  & \multicolumn{1}{r|}{208}  & 154 & 49 \\
31 & flights\_1     & \multicolumn{1}{c|}{10} & 560 & 80 & \multicolumn{1}{r|}{160} & 4680 & 667 & \multicolumn{1}{r|}{1379} & 168 & 10 \\
32 & services\_3    & \multicolumn{1}{c|}{5}  & 131 & 19 & \multicolumn{1}{r|}{38}  & 959  & 143 & \multicolumn{1}{r|}{290}  & 143 & 54 \\
33 & flights\_3     & \multicolumn{1}{c|}{8}  & 65  & 10 & \multicolumn{1}{r|}{19}  & 420  & 75  & \multicolumn{1}{r|}{116}  & 133 & 79 \\
34 & trains\_1      & \multicolumn{1}{c|}{7}  & 58  & 9  & \multicolumn{1}{r|}{17}  & 415  & 67  & \multicolumn{1}{r|}{117}  & 131 & 76 \\
35 & homes\_2       & \multicolumn{1}{c|}{8}  & 62  & 9  & \multicolumn{1}{r|}{18}  & 424  & 56  & \multicolumn{1}{r|}{139}  & 140 & 89 \\
36 & rentalcars\_2  & \multicolumn{1}{c|}{6}  & 77  & 11 & \multicolumn{1}{r|}{23}  & 631  & 91  & \multicolumn{1}{r|}{185}  & 157 & 61 \\
37 & restaurants\_1 & \multicolumn{1}{c|}{9}  & 256 & 37 & \multicolumn{1}{r|}{74}  & 2098 & 297 & \multicolumn{1}{r|}{581}  & 153 & 10 \\
38 & music\_1       & \multicolumn{1}{c|}{6}  & 68  & 10 & \multicolumn{1}{r|}{20}  & 468  & 73  & \multicolumn{1}{r|}{142}  & 118 & 61 \\
39 & hotels\_4      & \multicolumn{1}{c|}{7}  & 80  & 12 & \multicolumn{1}{r|}{23}  & 559  & 99  & \multicolumn{1}{r|}{141}  & 134 & 72 \\
40 & media\_2       & \multicolumn{1}{c|}{5}  & 32  & 4  & \multicolumn{1}{r|}{10}  & 215  & 29  & \multicolumn{1}{r|}{71}   & 112 & 59 \\
41 & hotels\_3      & \multicolumn{1}{c|}{6}  & 90  & 13 & \multicolumn{1}{r|}{26}  & 737  & 100 & \multicolumn{1}{r|}{193}  & 157 & 64 \\
42 & rentalcars\_3  & \multicolumn{1}{c|}{7}  & 44  & 7  & \multicolumn{1}{r|}{13}  & 332  & 55  & \multicolumn{1}{r|}{99}   & 148 & 72 \\
43 & hotels\_1      & \multicolumn{1}{c|}{7}  & 99  & 14 & \multicolumn{1}{r|}{29}  & 868  & 105 & \multicolumn{1}{r|}{250}  & 161 & 71 \\
44 & homes\_1       & \multicolumn{1}{c|}{7}  & 244 & 35 & \multicolumn{1}{r|}{70}  & 1829 & 282 & \multicolumn{1}{r|}{540}  & 159 & 81  \\ 
\bottomrule
\end{tabular}%
\caption{Statistics of the 15 services we used in experiments. 
}
\label{tab:dataset_count}
\end{table*}

\begin{table*}[t]
\centering
\begin{tabular}{c|ccccccccccccccc}
\toprule
Task order & \multicolumn{15}{c}{Tasks' IDs in order} \\
\midrule
Order1 & 30 & 31 & 32 & 33 & 34 & 35 & 36 & 37 & 38 & 39 & 40 & 41 & 42 & 43 & 44 \\
Order2 & 39 & 33 & 36 & 42 & 40 & 37 & 38 & 34 & 32 & 35 & 41 & 31 & 30 & 44 & 43 \\
Order3 & 30 & 41 & 38 & 31 & 43 & 39 & 40 & 33 & 34 & 44 & 37 & 36 & 32 & 35 & 42 \\
Order4 & 43 & 40 & 44 & 38 & 30 & 37 & 31 & 39 & 32 & 35 & 41 & 34 & 33 & 36 & 42 \\
Order5 & 30 & 33 & 44 & 31 & 38 & 32 & 42 & 40 & 37 & 43 & 36 & 39 & 41 & 35 & 34 \\ \bottomrule
\end{tabular}%
\caption{Five task orders of all our 15 tasks experiments.
}
\label{tab:dataset_order}
\end{table*}

\section{Implementation}
\label{sec:imp}
For reasoning generation by the teacher model~\cite{feng2021completing}, we utilized the following hyperparameters:

\begin{itemize}[leftmargin=*,itemsep=2pt,topsep=0pt,parsep=0pt]
\item \textbf{LLaMA-2 (70B)}: Model set as Llama-2-70B-chat-GPTQ, with temperature at 0.7, top\_p at 0.9, top\_k at 40, and a maximum of 512 new tokens.
\item \textbf{ChatGPT}: Generation was conducted from October 21th to 30th, 2023, using the `gpt-3.5-turbo' API. The settings were temperature = 0.7 and max\_tokens = 256.
\end{itemize}

For training the student models, we applied these hyperparameters:

\begin{itemize}[leftmargin=*,itemsep=2pt,topsep=0pt,parsep=0pt]
\item \textbf{T5-small (60M)} and \textbf{T5-base (220M)}: Training was conducted with a learning rate of 3e-4, batch size of 8, maximum input length of 512 (1024 for reasoning fine-tuning), maximum target length of 128 (512 for reasoning fine-tuning), and 5 epochs.
\item \textbf{T5-XL (3B)}: To optimize training time, we used LORA with a learning rate of 3e-4, batch size of 2 (8 for reasoning fine-tuning), maximum input length of 512 (1024 for reasoning fine-tuning), maximum target length of 128 (512 for reasoning fine-tuning), and 5 epochs. Lora settings included r = 8, alpha = 16, dropout = 0.05, targeting modules [ `q', `v' ]. For testing, we set max new tokens to 128 (512 with reasoning).
\item \textbf{LLaMA (7B)}: Also utilizing LORA for efficiency, with a learning rate of 3e-4, batch size of 128, a cutoff length of 512 (1024 for reasoning fine-tuning), and 5 epochs. Lora settings were r = 8, alpha = 16, dropout = 0.05, targeting modules [[q\_proj,k\_proj,v\_proj,o\_proj]]. For testing, settings included temperature = 0.02, top\_p = 0, top\_k = 1, num\_beams = 1, max new tokens = 128 (512 with reasoning).
\end{itemize}

Experiments are carried out using Nvidia RTX 3090 GPUs. Training durations for various models on total 15 tasks are as follows: T5-small takes about 1 hours on a single 3090 GPU, T5-base takes nearly 2 hours on a single 3090 GPU, FlanT5-XL takes about 16 hours on a single 3090 GPU, and LLaMa-7B takes 21 hours on two 3090 GPUs. Results are averaged across five different task orders and include the standard error in the tables and plots provided~\cite{feng2023towards2}.

\textbf{Details on Perturbation Operation}:
We altered the ground truth value with a random value from the dataset for the same slot for value-level perturbation. However, given the robustness of the teacher model, which can often detect errors and provide correct explanations even if the misleading value does not cause significant confusion, we introduced a slot-level perturbation. This involves completely changing the slot and substituting it with a random slot-value pair from the current task.

To address the ``Value Selection Quandary'' common in long dialogues and to save the time and cost of using teacher LLMs, we engage teacher models to generate reasoning for samples with turn id greater than 10, aligning with findings from Figure \ref{fig:motivation}(a). For briefer dialogues, where typically only one possible value appears, our examination indicates teacher models usually return direct reasoning, like ``In the given dialogue, the user explicitly mentions their destination as <1150 Webster street>. Therefore, the answer to the slot <ridesharing-destination> is 1150 Webster street.''. Thus, for uniformity across all short dialogues, we've standardized the prompt to ``In the given dialogue, the value of the requested slot is explicitly mentioned,'' enabling student models to learn from this consistent template.

\section{Time Complexity Comparison}
Firstly, we examined the training time of the student model. When using LLaMA-7B as the backbone, the inclusion of additional rationales led to an approximately 1.5 times increase in training time compared to the original training. For smaller student models like T5-small, the training time remained consistent with no inclusion of reasoning.
In terms of inference speed, a comparison is presented in the table \ref{tbl:time} below:
\begin{table}[h]
\centering
\scalebox{0.65}{
\begin{tabular}{lrrrr}
\toprule
Inference Speed \\ (Samples/Min) & T5-small & T5-base & FlanT5-XL & LLaMA-7B  \\
\midrule

\rule{0pt}{4pt}Vanilla Method	   &857 &222 &51 &41	 \\
\rule{0pt}{8pt}RoS (ours)  &86 &24 &7 &5 \\

\bottomrule
\end{tabular}}
\caption{Inference time for different models.}
\label{tbl:time}
\end{table}
The table \ref{tbl:time} illustrates a ninefold increase in time consumption due to generating rationales. For instance, during the generation of rationales, the LLaMA-7B experiences a decrease in samples per minute from 41 to 5. Despite the reduction in inference speed, this change contributes to a significant performance improvement.